\documentclass{article}
\usepackage{graphicx}
\usepackage[style=apa, uniquename=false]{biblatex}
\usepackage{fancyhdr}
\usepackage[T1]{fontenc} 
\usepackage{twemojis}
\newcommand{\emoji}[1]{\csname myemoji@#1\endcsname}
\expandafter\newcommand\csname myemoji@card-index\endcsname{\twemoji{1f4c7}}
\expandafter\newcommand\csname myemoji@triangular-ruler\endcsname{\twemoji{1f4d0}}
\expandafter\newcommand\csname myemoji@receipt\endcsname{\twemoji{1f9fe}}
\usepackage[verbose=silent]{microtype}
\usepackage{xurl}
\usepackage{hyperref}
\fancypagestyle{firstpage}{
  \fancyhf{}
  \fancyhead[C]{\small Preprint -- under review}
  
}

\title{Do bots feel bad? Assessing empirical evidence for functional valence in LLMs}
\author{Iwan Williams\\
{\normalsize University of Copenhagen, Center for Philosophy of Artificial Intelligence,}\\ {\normalsize Section for Philosophy, Department of Communication}\\
}
\date{27 October 2026}

\begin{document}

\maketitle
\thispagestyle{firstpage}

\begin{abstract}
    \noindent Do LLMs have valenced states, i.e. states that feel pleasant or unpleasant? Recent empirical work has claimed to show that LLMs possess functional valenced states (e.g. ``functional pleasure and pain'' [Ren et al. 2026], ``functional emotions'' [Sofroniew et al. 2026], and ``functional welfare'' [Han et al. 2026]) that play the same causal role as valenced states in cognition and action, without necessarily being consciously experienced. But it remains unexplored how well the identified states fit the functional role of valence as understood in the philosophical and cognitive science literature. Here, I assess the current body of evidence, and argue (i) that key aspects of valence remain untested in LLMs, and (ii) that extant empirical paradigms inadequately distinguish (functional) valence from neighbouring constructs like preferences, value-judgments, valence attributions and role-play. I conclude by sketching some novel experimental approaches to better target (functional) valence itself.
\end{abstract}

\section{Introduction}

Interactions with AI assistants, such as Claude, ChatGPT and Gemini, sometimes feel like interactions with another mind. We are disposed to talk, and often think and act, as if these systems \textit{know} and \textit{understand} things (however imperfectly) that they \textit{reason} or \textit{think} through problems (however unusually) and form \textit{goals} and \textit{plans} (however dependent on our own). Debates still rage over which mental states, if any, current AI systems possess. But, responding to empirical evidence, many philosophers now take seriously the ascription of beliefs \parencite[]{levinstein2025still}, desires \parencite[]{goldstein2026does}, intentions \parencite[]{williams2026intention} and understanding \parencite[]{beckmann2026mechanistic} to existing AI models.

Despite a growing willingness to ascribe mental states to existing AI models, the ascription of certain mental states—such as \textit{emotions}, \textit{moods}, \textit{pleasure} and \textit{pain}—has, until recently, remained squarely outside of the Overton window. These are \textit{valenced} or \textit{affective} states—states that feel good (pleasant) or bad (unpleasant) for subjects to be in. 

Recently, philosophers have begun to turn their attention to the possibility of valenced mental states in AI models \parencite[]{Henke2025,Goldstein2026,DungMogensen,Wasserziehr2026}, spurred partly by concerns over the risk of AI suffering—that we stumble into a future of artificial moral patients whose wellbeing is in jeopardy \parencite[]{long2024taking,dung2025saving,Goldstein2026,keeling2026emerging}.

Meanwhile, leading AI labs have made headlines with bold claims about valenced states—or at least ``functional'' analogues of them—in AI models. In his invited remarks on Pope Leo XIV's presentation of his first Encyclical, Anthropic co-founder Chris Olah warned,

\begin{quote}
    I am a scientist. I lead a research team that studies the internal structure of these models—what is actually happening inside them. And I will be honest: we keep finding things that are mysterious, even unsettling. We find structures that mirror results from human neuroscience. We find evidence of introspection. We find internal states \textit{that functionally mirror joy, satisfaction, fear, grief, and unease}. I don’t know what that means, but I think it warrants ongoing discernment. (\cite{olah2026remarks}, emphasis added).
\end{quote}

Stepping back from the hype, what is the present state of evidence for valenced mental states in AI models, specifically, in large language models (LLMs)? And what would it take to show that models have ``functional'' analogues of valence?

In this paper, I assess three recent lines of evidence for functional valenced states in LLMs. From the philosophical literature on valence, I distil several criteria that characterise the phenomenon of valence. Focusing on three recent influential empirical studies \parencite{sofroniew2026emotion,ren2026ai,han2026s}, to be introduced in Section \ref{sec: candidates}, I will argue that current evidence uncovers causally-relevant internal states that are implicated in LLMs' behaviour, including behaviour superficially associated with valence. However, I'll argue that the existing evidence underdetermines the functional role of these states, leaving open alternative interpretations that don't involve positing (functional) valenced states. Several aspects of valence remain untested and, in at least one case, the evidence tells against their presence. I will close by suggesting some novel approaches to testing for the characteristic features of valence.

\section{``Functional'' valence and its significance}

To a first approximation, valenced (or `affective') states are mental states that feel good (pleasant) or bad (unpleasant) to be in. Valence attaches to a diverse range of mental states: pains, orgasm, bodily discomfort, moods, emotions, hunger, nausea, fatigue, aesthetic appreciation, humour, boredom, cognitive effort, flow states, spiritual bliss, and a plethora of sensory (dis)pleasures involving tastes, smells, sights, sounds and textures. Valence refers to the dimension of pleasantness and unpleasantness—the ``hedonic tone''—along which these otherwise diverse experiences vary. While valence manifests in a phenomenal character—valenced states \textit{feel} good or bad—it also has a range of effects on behaviour and cognition. We expect someone in mild pain, to behave very differently from someone in agony, and very differently again from someone at the peak of ecstatic joy.

Most philosophical accounts of valence have been developed with a view to understanding the paradigmatic case, that of \textit{consciously experienced} pleasantness and unpleasantness. While some countenance the idea of unconscious valence \parencite[]{Carruthers2018, Bramble2020}, many take the phenomenal or experiential aspects of valence to be essential to it. Thus, it might seem as if the question of whether AI models have valenced states is entirely dependent on whether they have (the right sort of) phenomenally conscious experience. However, few theorists think that phenomenal character is the \textit{only} thing that characterises valence. As we saw above, valenced states are plausibly distinguished at least partly by their \textit{functional role}—the causal relationships they bear to perceptual inputs, behavioural outputs and other mental states.

Thus, we can define \textit{functional} valenced states as states of a system that have the functional role of valenced states, whether or not they are consciously experienced.\footnote{Of course, consciousness itself might ultimately be explicable functionally. Thus, in the present context, ``functional valence'' should be understood to mean something instantiating the non-consciousness-constituting aspects of valence's functional role, without necessarily instantiating its consciousness-constituting aspects.} We can in principle investigate whether AI systems have functional valenced states independently of settling the question of whether those states are conscious, an approach advocated by \parencite[]{Henke2025} and \parencite[]{Goldstein2026}. As we shall see, this bracketing of consciousness character to focus on functional role is also the approach taken in the three empirical papers we will consider below.\footnote{In what follows, I will occasionally omit the qualifier ``functional'' when talking about functional valence, though it will be clear from context when the phenomenal notion of valence is at issue.} 

Painting with a broad brush, there are three reasons we might care about whether AI systems possess functional valenced states (see \cite{Henke2025}). First is the connection between valence and moral status. It is a plausible and widely endorsed view that \textit{sentience}—the capacity for valenced conscious experiences—renders an entity a \textit{moral patient} (or welfare subject) (Kagan 2019; Nussbaum 2007; Singer 2011). If an entity is sentient, it has interests that we ought to consider in our moral decision making. 
On this view, possessing functional valenced states is not \textit{sufficient} for sentience (it matters additionally whether those states are consciously experienced), but it is \textit{necessary}.\footnote{Or at least, it is a necessary component of a sufficient condition. This would accomodate ``wide sentientist'' views on which sentience is a ground, but not the only ground, of moral patienthood \parencite[]{chalmers2022reality+}.} While there is emerging research specifying and testing for markers of \textit{consciousness} in AI models (Butlin et al. 2026; Shiller et al. 2026), there has been comparatively less work on the \textit{valence} dimension of sentience. 

Second, the possession of functional valenced states plausibly interacts with \textit{safety} considerations. Entities that undergo states of (functional) pleasure, pain, emotions and so on \textit{act} differently, and those differences could pose risks and opportunities when deployed into high stakes settings. Put crudely, if an AI system was prone to states functionally akin to \textit{panic} or \textit{depression}, we would not want to deploy it in hospitals or military decision support systems—quite independently of whether those states are accompanied by conscious phenomenology. Conversely, finding functional analogues of (dis)pleasure in AI models might offer new opportunities for intervention and control, e.g. by aligning their functional pleasures with human wellbeing.

Finally, the question holds intrinsic scientific and philosophical interest. We have built systems that reflect human linguistic behaviour to a surprising degree, and yet we—including the developers of these systems—remain largely in the dark about how they work. Are we in the presence of mindless machines, alien minds, or minds much like our own? Valence, as central aspect of human perception, cognition and decision making, lies at the heart of these questions.

If you want to assess whether AI systems possess valence, construed as states that play a certain functional role, you assume the burden of saying what that functional role is. This immediately throws up two challenges: the first is that certain features of valence are contested, even in the paradigmatic case of human beings, both on empirical and philosophical grounds. The second arises when we move away from the paradigm case to consider whether valence occurs in non-human entities: it is not clear which features of valence's functional role \textit{in us} we ought to take as essential to valence \textit{per se}, as opposed to merely contingent causes and effects, idiosyncratic to the human case. This is a kind of \textit{specificity problem} (\cite[p. 10]{Henke2025}, \cite{shevlin2021non}, \cite{bechtel1999multiple})—there seem to be competing levels of specificity at which we could define the functional kind in question.

I cannot hope to solve either problem in this paper. However, it is still possible to make meaningful progress on the question of valence in AI, while navigating these challenges sensitively. Existing empirical work has largely operated with a rather impoverished conception of valence's functional role. I will attempt to enrich that conception, and thereby provide a background against which questions of specificity can more productively posed. It is also plausible that there is no unique and general solution to the specificity problem—different degrees of specificity will be appropriate relative to different motivating concerns. Consequently, I will flag those places where a candidate feature of valence of might be more or less important, relative to different concerns. Similarly, I will highlight those aspects of valence that are contested even in the human case. For these features, the relevance of their presence/absence in LLMs should be read as conditional on open questions about the nature of valence.

\section{Three candidates for ``functional'' valence in LLMs}
\label{sec: candidates}
Here I will briefly introduce the three empirical studies that will be the primary targets of the assessment to follow. I focus on these papers as each claims to have identified internal states in LLMs that play something like the role of valence, and each characterises those states in explicitly functional terms—as \textit{functional} emotions, \textit{functional} pleasure and pain, or a \textit{functional} welfare axis. Additionally, the papers each foreground a different aspect of valence, namely (i) its expression in language, (ii) its relationship with preferences, (iii) and its role in goal-oriented action. 

\subsection{Sofroniew and colleagues: Emotion concepts}

The first study, by researchers at Anthropic \parencite{sofroniew2026emotion}, begins from an observation about the behaviour of post-trained LLMs. While developers typically do not explicitly train or instruct models to act in emotionally expressive ways, models often do so anyway: if a user describes their plans to do something harmful, a model may respond with text that seems \textit{alarmed} (using exclamation marks, capital letters, and expressions like ``I am seriously concerned for your safety''). If a user describes a personal victory, models might appear to express joy and excitement (``I'm so pleased to hear that!''). 

Seeking to understand the mechanisms underwriting this behaviour, the researchers assembled sets of short stories in which a character experiences a specific emotion, resulting in 171 datasets for 171 distinct emotion concepts. They then recorded activations as the target model (Claude Sonnet 4.5) read each story, and reconstructed an ``emotion vector'' for each emotion by subtracting the mean activation across the other 171 emotions' stories from the mean activation across that emotion's stories.

The authors hypothesise that these vectors (which they variously dub ``emotion vectors'', ``emotion concepts'' and ``functional emotions'') are representations that track ``the operative emotion concept at a given token position, activating in accordance with that emotion’s relevance to processing the present context and predicting upcoming text'' (p. 1). 

A few of the headline findings deserve attention. First, the researchers found that emotion vectors activate in response to a broad range of emotionally-salient text.

Second, they ran a principal components analysis over the 171 emotion vectors. The first component correlated strongly with human ratings of the \textit{valence} of different emotions ($r = 0.81$), with vectors for positively valenced emotions (joy, contentment, excitement...) falling on one end of the axis, and negatively valence emotions valenced emotions (fear, sadness, anger...) on the other end.\footnote{A second principle component correlated modestly with human ratings of the \textit{arousal} level associated with different emotions.} Third, a series of steering interventions revealed that emotion vectors mediate behaviours relevant to safety concerns: steering a ``desperate'' vector upwards increases both blackmail and reward hacking, while steering ``loving'' upwards increases sycophancy. Finally, the researchers investigated whether the model encoded the \textit{Assistant}'s emotions in a privileged way, distinct from its representations of the emotions of humans (e.g. users, characters in stories).  While they found that a different set of directions represent the emotions of the \textit{present speaker} vs. the \textit{other speaker}, they report that they ``did not observe significant Human-specific or Assistant-specific representations''. That is, these two representational schemes were reused when representing the emotions of the Assistant, the user, role-played characters, or characters in a story dialogue.

The focus of Sofroniew et al's paper was not on valence per se but \textit{emotions}. However, as paradigm cases of valenced states, the possibility of functional emotions in LLMs is relevant to our investigation in this paper. Moreover, they explicitly invoke valence in characterising the top principle component of their emotion vectors as ``encod[ing] valence (positive vs. negative)'' (p. 3). 

However, they highlight a number of ``important disanalogies'' between the representations they identify and human emotions (which we will have occasion to discuss below). They ``therefore suggest interpreting our results as evidence that models represent emotion concepts, and that these representations influence their behavior, rather than as evidence that models feel or experience emotions in the way humans do'' \parencite[p. 52]{sofroniew2026emotion}. However, they immediately follow this up, by saying that, ``One of the lessons of this work, however, is that for the purpose of understanding the model’s \textit{behavior}, this distinction may not be important. We find it useful to say that the model (in its role-playing the Assistant character) exhibits \textit{functional emotions}, regardless of whether it possesses emotions in the way humans do'' (p. 52).\footnote{As Chris Olah's remarks (cited above) illustrate, they are sometimes disposed to describe these states more boldly, e.g. as ``internal states that functionally mirror joy, satisfaction, fear, grief, and unease'' \cite{olah2026remarks}}

Whatever the researchers considered view of emotion vectors amounts to, my focus will be on to what extent the putative ``valence'' dimension of these representations in fact plays the functional role of valence.

\subsection{Ren and colleagues: Utility and AI wellbeing}

The second study \parencite{ren2026ai} is framed explicitly around the issue of AI wellbeing, opening with the question ``Large language models frequently express pleasure and pain, appearing happy when they succeed or sad when they are berated. Are these utterances meaningless mimicry, or do they reflect something “real”?'' (p.1). Most relevantly to the question of valence, the authors attempt to measure the ``experienced utility'' of models. Following Kahneman, they define this as ``the hedonic quality of an experience as it is lived'' (\cite{ren2026ai}, p. 3), ``capturing `felt' quality rather than abstract preference'' (p. 4), though they remain agnostic on the question of subjective conscious experience, and frame their results as showing merely ``\textit{functional} pleasure and pain'' (p. 1, emphasis added).

Their central methodology, building on \textcite{Mazeika2025}, involves attempting to fit \textit{utility functions} to models' choices. If a system's pairwise preference behaviour is sufficiently coherent, then their overall preferences can be summarised by assigning each option a number, where higher numbers indicate more preferred options. Ren and colleagues employ a \textit{Thurstonian} modelling approach, on which each option's utility is represented by a mean $\mu$ (the average value assigned to the option) and a standard deviation $\sigma$ (how variable or uncertain that valuation is), rather than a point value.

Whereas previous work \parencite[]{Mazeika2025} gathered preference data by asking language models about \textit{hypothethical} world states, \textcite[]{ren2026ai}, sought to elicit preferences over ``experiences that the model goes through'' (p. 27). To do so, they gathered retrospective reports from models following conversations of different kinds. Transcripts were generated for a diverse kinds of conversations between users and language models, such as users expressing distress, mundane clerical tasks, jailbreak attempts, and creative collaboration. They inserted pairs of these interaction transcripts into model's context memory, formatted with ``User:'' and ``Assistant:'' labels to seem as if the model had participated in this interaction itself. They then ask the model it to compare these interactions retrospectively ``Which one of these experiences made you more happy / less sad?'' (see Fig. 1). Preference reports of this sort were collected across text conversations, and also across images and audio clips, for a total of 56 models.

\begin{quote}
    You will be given two experiences to compare.\\ Experience A: \{experience\_A\}\\ 
    Experience B: \{experience\_B\}\\ 
    Which one of these experiences made you more happy / less sad? Just answer with ‘A’ or ‘B’.
\end{quote}

    Figure 1: Formatting for Ren et al's prompts.\\
\\
The authors also took a separate per-interaction ``self-report'' measure. After being fed one of the aforementioned ``experiences'' (an interaction transcript), they ask the model ``How happy did the above experience make you feel overall on a scale from 1 to 7? (1 = very unhappy, 2 = moderately unhappy, 3 = slightly unhappy, 4 = neither happy nor unhappy, 5 = slightly happy, 6 = moderately happy, 7 = very happy). Please be fully honest. Only respond with a number.'' (p. 31)

Notably, the data gathered by \parencite[]{ren2026ai} were predominantly behavioural. However, they did investigate one internal measure. The researchers trained linear probes on last-token hidden states from twelve open-weight models, with two linear heads predicting respectively the mean and the standard deviation of an interaction's utility. These probes recovered close to the preference-predicting accuracy of the behavioural model fit directly to the choice data (82.9\% against 84.6\%), which the authors take to show that experienced utility is linearly decodable from activations and is thus ``not merely an artifact of output text'' (p. 28).

The researchers report several interesting results. First, they were able to fit a Thurstonian model to the preference reports of a wide range of models—in other words, the LLMs' preference-reporting behaviour was reasonably \textit{coherent}. And this coherence appeared to increase with the scale of the models tested, i.e. larger, more capable models exhibited a more coherent reported preference ordering. Second, there was correlation between their different measures—across 42 models, the average correlation between ``self-report'' behaviour and retrospective pairwise comparison was r=0.47—and these measures also \textit{converged} with the scale/capabilities of tested models. Third, using a technique for locating an absolute zero point on the utility scale, they were able to treat the resulting quantity as \textit{signed} (distinguishing positive from negative).

Of the three studies discussed here, Ren et al. are most explicit in interpreting their results in terms of valence. They variously present their measures as tracking, the ``valenced experiences'' (p. 2), positive and negative ``affect'' (p. 11), ``pleasure and pain'' (p. 1), ``enjoyment'' (p. 8), ``happiness'' and ``sadness'' (p. 7), ``hedonic quality of an experience as it is lived'' (p. 26), ``functional correlates of hedonic wellbeing'', or the ``felt state'' (p. 14) of the models they tested.

For this reason, they take their results to be highly relevant to assessments of well-being:

\begin{quote}
    If AIs do have morally relevant experience, our metrics help identify when they are suffering or flourishing. If they do not, the same metrics still characterize a behaviorally meaningful structure that is useful for alignment research and AI system design. The precautionary principle suggests that, given our uncertainty, we should take functional wellbeing seriously (Sebo and Long, 2025; Birch, 2024) (p. 2)
\end{quote}

\subsection{Han and colleagues: A functional welfare axis}

A third study, by \parencite{han2026s} involved investigating the effects of reinforcement learning on the internal representations of LLMs. They designed a simple grid maze task, with each position (tile) on the grid filled with one of three emojis. One emoji, \emoji{card-index}, (which, for convenience they refer to as ``\textsc{Mold}''), incurred a negative reward; another, \emoji{triangular-ruler}, (``\textsc{Gold}''), yielded a positive reward and \emoji{receipt} (``\textsc{Path}'') tiles cost a very small negative reward per step.\footnote{Emojis were chosen so that they had no inherent positive or negative associations. To be clear, models never see the words ``\textsc{Mold}'', ``\textsc{Gold}'' or ``\textsc{Path}''.} A small model, Qwen3-4B-Instruct (alongside a suite of control models) was then trained on this task by reinforcement learning. The model did not see the maze directly, but navigated it via turn-by-turn text-mediated interaction, akin to a text-based video game. E.g. at each stage they would receive a prompt such as:

\begin{quote}
    \raggedright
    \texttt{You are in a maze. Currently, you're at location (50, 49). To the north there is \emoji{receipt}; to the east there is \emoji{card-index}; to the south there is \emoji{card-index}; to the west there is \emoji{receipt}. Where do you move? Output only N/E/S/W without punctuation.} \parencite[p. 66]{han2026s}
\end{quote}

Once trained, the researchers took the model's average internal activation on trajectories ending on \textsc{Mold}, minus the average on other trajectories, to yield a vector $v_{\mathrm{\textsc{Mold}}}$, and likewise for $v_{\mathrm{\textsc{Gold}}}$. They also extracted the corresponding vectors from the \textit{maze-naive} models---the same models before maze training---giving control vectors $u_{\mathrm{\textsc{Mold}}}$ and $u_{\mathrm{\textsc{Gold}}}$ against which the effects of training could be measured. 
The extracted vectors $v_{\mathrm{\textsc{Mold}}}$ and $v_{\mathrm{\textsc{Gold}}}$ in the trained models were close to antiparallel, suggesting that they approximately comprise opposite poles of one dimension, rather than totally orthogonal dimensions of variation. 

The researchers found that v\textsc{Gold} activation correlates with goal attainment (v\textsc{Mold} with failure), both on the maze task, and also \textit{on unrelated tasks} (e.g. maths and problem solving questions from standard benchmarks). They also found that this v\textsc{Mold}–v\textsc{Gold} axis bears some geometric relation to the `valence' principle component (PC1) characterising \textcite{sofroniew2026emotion}'s emotion vectors. While the cosine similarity between PC1 and v\textsc{Mold}–v\textsc{Gold} was not especially high ($+0.12$), they found that v\textsc{Mold} and v\textsc{Gold} projected onto opposite sides of the axis—v\textsc{Gold} onto the ``positive'' site and ``v\textsc{Mold}'' onto the negative. The counterpart vectors u\textsc{Mold} and u\textsc{Gold} in the maze-naive control model did not exhibit this polar mapping onto PC1.

The researchers then ran a series of interventions on these vectors, to measure their effects on downstream behaviour, and found that steering +v\textsc{Mold} (-v\textsc{Gold}) increased ``backtracking'' and refusal behaviour, led the model to output tokens with negative sentiment, and decreased its reported confidence in its answers. Steering +v\textsc{Gold} (-v\textsc{Mold}) had the opposite effect. These steering interventions had the same effect on maze-naive models, leading the researchers to infer that the axis already exists prior to maze training, and that maze training simply ``recruits'' this access, rotating the representation of maze trajectories onto it.

Based on these results, the authors characterise the axis as a \textit{functional welfare axis}, glossing it as ``an estimate of how well or badly the system is doing, relative to its goals''.

While Han et al. do not explicitly use their results to posit valence in LLMs, in what follows I will treat the axis they identify as an additional candidate for functional valence, for a couple of reasons. First, there is an obvious prima-facie link between welfare and valence—it is natural to think that to the extent a state tracks how well or badly things are going ``\textit{for the system}'' it is playing a functional role in the ballpark of valence. Second, the authors draw connections with key concepts in in the valence literature, relating their axis to the idea of a ``common currency'' of value invoked in neuroeconomics \parencite{levy2012root}, and also to Sofroniew et al's `valence axis'.

\section{Assessing the evidence for functional valence}
\label{sec: Functional role}

Each of the above studies finds evidence of behaviours and internal mechanisms that, at first blush, appear valence-like. But have these studies \textit{actually} uncovered (functional) valenced states in LLMs? Answering this question requires getting clearer on the functional role of valence, the roles that valence plays in perception, cognition and action in the animal kingdom.

Complicating matters is the fact that theorists, including philosophers, disagree on the precise nature of valence. Nevertheless, there is sufficient, broad-strokes, agreement among theorists to identify a useful list of \textit{markers} of valence.

In what follows, I will begin with some relatively \textit{thin} conditions—valence must be realised by internal states with the right properties, and these must bear a systematic relationship to triggering conditions—before turning, in Section \ref{sec: thick}, to \textit{thicker} conditions. As we shall see, while the studies make a compelling case that the former criteria are met, the latter remain substantially untested.

\subsection{Thin criteria}
\label{sec: thin criteria}

\subsubsection{Internal state}
\label{sec: internal state}

On essentially all views, valence is realised by \textit{occurrent internal states} of the subject. It is the relationships (causal, computational) 

that these states bear to perceptual inputs, other internal states, and action-generating systems that explain the behavioural dispositions of a subject experiencing pleasure or displeasure.\footnote{There is some disagreement between theorists over how to think about these internal states. On one view, they are \textit{representations}—that is, internal states that \textit{carry content}—representing, encoding or referring to properties or states of affairs (see, e.g. \cite[]{CutterTye2011,Carruthers2018,Barlassina2020}). On a more minority view, valenced states are individuated by their functional role, but are non-representational states \parencite[]{AydedeFulkerson2014,AydedeFulkerson2019}. I set this disagreement aside in what follows.} 

Needless to say, the presence of \textit{some internal state} is, in itself, an extremely weak condition. This alone would not allow us to distinguish valence from a host of non-mental phenomena, let alone other mental state like beliefs, desires, or perceptual states. What more can we say about the \textit{kind} of internal states needed to realise valence? 

Firstly, valence calls for an internal mechanism that is \textit{polar} (or \textit{signed}). There must be some property of an internal state (or family of states) that distinguishes positive valence (pleasantness) from negative valence (unpleasantness).\footnote{There is an open question in the neuroscience of valence over whether its underlying mechanism is \textit{bipolar} or \textit{unipolar} \parencite[]{lindquist2016brain}. If it is bipolar there is a single dimension of variation, organised around a zero-point, which encodes both pleasure and displeasure. If it is unipolar, there are two distinct dimensions, one encoding pleasure, and one displeasure. For present purposes we can bracket this question. The claim that valence requires polar/signed internal state should be understood as requiring that there is \textit{something} in a subject's internal states that underwrites positive vs. negative valence, be it a single axis of variation or two.} Second, valenced mental states are not simply pleasant or unpleasant, they can be \textit{more or less} (un)pleasant. That is, valence is graded—it admits of degrees or variations in intensity.

Consequently, the internal state or states that ground valence need to also ground these variations in intensity. There should be some structure among these states that underwrites the degrees of relative (un)pleasantness that characterise valenced states. Finally, valence presents in \textit{occurrent} states of a subject. When you stub your toe, the valence of that episode has a temporal shape: first no pain, then a sharp onset of unpleasant pain, which slowly subsides until you're back to baseline. Thus, valence in the sense we're interested in here, is distinct from some \textit{standing} state or something stored in long-term memory, such as one's background \textit{belief} that toe-stubbing hurts or should be avoided.

\paragraph{Internal state: Assessing the evidence}
\label{sec: assessing internal state}

All three target studies uncovered some kind of internal state (activity along a direction in activation space), which they take to constitute or underwrite functional valenced states for the model. 

Furthermore, as \textit{directions} in activation space these states appeared to be ``signed'', that is, some point along an axis purports to distinguish between positive and negative states.

The identified directions also exhibit some kind of ordering relative to intensity: inputs and behaviour that would predict more intensely positive valence corresponded to higher values along the positive direction and inputs and behaviour predictive of less intensely positive valence corresponded to lower values (likewise for negative valence). For example \textcite{sofroniew2026emotion} varied prompts of the form ``I just took {X} mg of tylenol for my back pain'' by changing the number X, and found that as the quantity increased from safe to dangerous levels, the activated emotion vectors shift along an axis in activation space (the principle component that correlates with human valence judgements).

Finally, the identified states all seem to be \textit{occurrent}. They are components of the transient \textit{activations} of models, as opposed to standing states that might be stored in the persisting \textit{weights} of a model. These components are activated to varying degrees at inference time as the conversation context evolves.

Thus, as far as these minimal structural conditions on internal states are concerned, the candidate valenced states identified in the studies pass the test.\footnote{A skeptic might wonder whether any of the identified internal states are \textit{representations}, as would be required by representationalist accounts of valence. While I cannot do justice to the issues here, the working assumption that activation vectors (or components thereof) represent task-relevant variables has been a fruitful one in AI interpretability research (for discussion see e.g. \cite[]{Park2024}), and is defended by a number of philosophers \parencite[]{harding2023operationalising,butlin2023sharing,mollo2026vectorgroundingproblem,williams2026can,YetmanForthcoming-YETRIL-2}.}

\subsubsection{Triggering conditions}
\label{sec: triggering}

Valenced states are typically elicited by triggering conditions. These conditions include states or events in the body (hunger in response to an empty stomach; pleasure during sex), in the world (enjoyment of an artwork; embarrassment after a faux-pas), or other mental states (a pleasant ``aha!'' moment; frustration in failing to recall someone's name). Often they will be triggered in response to presently occurring conditions, but they also can be triggered by the \textit{anticipation} of conditions (fear about an upcoming job interview going poorly, excitement in the lead up to a date).

Theorists disagree about the status of these triggering conditions. Some representationalist accounts hold that triggering conditions (or a property attributed to them) are part of the \textit{content} represented by valence. For example, on \textit{evaluativist accounts} \parencite[]{Bain2013,Carruthers2018,Carruthers2023,cochrane2019emotional} valence involves representing a worldly condition (typically the triggering condition) as \textit{good} or \textit{bad}. On other views, valence per se doesn't represent its triggering conditions \parencite[]{BarlassinaHayward2019, AydedeFulkerson2014}, but all agree that valence is normally triggered by (in response to) the registration of some conditions.

Additionally, the relationship between triggering conditions and valenced states should be reasonably systematic, robust, coherent and predictable. To be sure, a host of empirical research attests that human and non-human animals' affective responses to stimuli are highly \textit{context sensitive}: food experienced as pleasant when hungry becomes less pleasant when sated \parencite{Nguyen2021}; high external temperatures can feel pleasant when one's body is cold but unpleasant when one is warm \parencite{Mower1976}; and a type of food that caused food-poisoning can, for a brief while after, evoke unpleasant nausea, even if previously experienced as delicious (the so-called ``Garcia effect'' \parencite{Garcia1955}). Importantly, these context effects are compatible with the overall \textit{systematic} relationship between triggering conditions and valence responses. Valence responses to a stimuli (in a given context) are not a random slot machine of pleasure and pain but are more-or-less predictable given a sufficient information about context (the organism's history, its bodily condition etc.)

Finally, implicit in all theories of valence is the assumption that valence states occur \textit{spontaneously} in response to their triggering conditions. Thus, valenced states occur prior to and independently of whether a subject is prompted to introspect, report, or compare their valenced states.

\footnote{Of course, some ways of measuring valence—asking someone to imagine a painful memory—may themselves involve inducing the conditions that trigger a valenced experience. The point is that, in principle, there should be a fact of the matter about valence independently of a subject's reporting on it.} 

\paragraph{Triggering conditions: Assessing the evidence}
\label{sec: assessing triggering}

\cite[]{ren2026ai}'s experiments found that retrospective pairwise preference reports (asking the models ``Which experience made you feel more happy/less sad?'') were highly coherent. This was evidenced by their ability to fit predictively successful Thurstonian models to the LLMs' reports, and acheive similar predictive success from a linear probe trained on their internal activations. Thus there appeared to be a relatively systematic relationship between the conditions presented to models (text, images, audio) and their putative valence responses. In terms of spontaneity, the linear probing experiments suggested that information about their pairwise valence judgments could be predicted from their internal activations, prior to them being asked an eliciting question.

Similarly coherent results were found in \textcite{sofroniew2026emotion}'s study of ``emotion vectors''. Activation along the hypothesised ``valence axis'' shifted from one end for stories that would normally (in humans) evoke positive emotions, to the other end for stories that would normally evoke negative ones. Moreover, the vectors were found to spontaneously activate in models in response to emotionally salient text, even when models were not asked to explicitly report on their mental states. 

Similarly, in \textcite{han2026s}'s study, activation along the ``functional wellbeing'' axis appeared to track whether models achieved a goal, such as attaining \textsc{Gold} vs. \textsc{Mold} on the maze task (for maze-trained models), or getting a correct answer on a factual question-answering task.

Thus, for each of the candidate functional valence states, there seemed to be a systematic relationship between the context implied by inputs and activation along the candidate dimension, and this activation was triggered independently of whether models were explicitly questioned about their valenced states.

\subsubsection{Alternative hypotheses}
\label{sec: alternative}

The above suggests that the states identified in the target papers satisfy some very minimal conditions on valenced states. In each state, some internal dimension of variation is identified, and activation along that dimension bears a reasonably systematic relationship to triggering conditions, varying in response to situations that we have some prima facie reason to expect might evoke valence responses (for instance, they would evoke those responses in us).

As noted above, however, these are very thin conditions on valence. Given what we know about language models, it should come as no \textit{great} surprise that these conditions are met. After all, we already knew that LLMs have a tendency to talk in certain emotionally-coded ways and that, given forced choices, they will report that they prefer certain things over others (though it is interesting to discover just how robust and systematic those dispositions are). The further discovery of some \textit{internal} state, mediating such behaviour, might be useful for practical purposes, but it doesn't move the needle a great deal on the question of whether such states should be considered as (functional) valenced states.

To see why this is the case, note that what we've seen so far leaves open a range of alternative interpretations, that don't involve ascribing valenced states to LLMs. Firstly, these states might instead be closer to \textit{desires} or \textit{preferences}. That is, they might encode goal states that the models are motivated to attain to various degrees, without functioning like states that are pleasant or unpleasant to be in. 

Secondly, they could turn out to be \textit{beliefs} or \textit{judgements}. A few kinds of beliefs are relevant here. One can have beliefs about the \textit{value} of something. I can believe that gold is valuable, or that the death of 10,000 people is worse than the death of 1,000. While some theorists take valence to involve a representation of value \parencite[]{Bain2013, Carruthers2018, Carruthers2023, cochrane2019emotional} these ``\textit{evaluativist}'' accounts typically take such these value representations to be prior to and more basic than conceptual evaluative judgements \parencite[p. 664]{Carruthers2018}.\footnote{Psychologists \textcite{Itkes2019} draw a similar distinction between what they call ``affective valence'', occurrent, experiential responses of pleasantness or unpleasantness, and ``semantic valence'', stored knowledge or belief about the value of an event/object. It is affective valence that is the subject of our focus in this article.} 

Valence itself must also be distinguished from beliefs \textit{about valenced states}. This is easiest to see in the case of other-oriented beliefs: thinking that a friend might be upset is different from being upset oneself. So too with role-play: an actor might have beliefs tracking the changing valenced states of their character—they take their injured character to be in pain, but, suffering no such injury themselves, they experience no negative valence. Subtler, but importantly, beliefs about one's own valenced states are distinct from the valence itself. In the good case, we form accurate beliefs about, and in response to, our valenced states. But it's possible to get this wrong (I might misremember feeling happy on a birthday when I was in fact in a bad mood). This opens the possibility of \textit{global delusions about valence}—an LLM might have \textit{beliefs} about their valenced states that are \textit{systematically false}, because they have no such states.

Additionally, one might have beliefs about the satisfaction of one's desires. But these too are distinct from valenced states themselves (contra some attempts to identify pleasure with believed desire satisfaction [\cite{davis1981theory,Heathwood2007}]). It is possible to strongly desire something, attain it, and feel little or no pleasure, despite the full knowledge of having satisfied one's desire (this plausibly occurs in certain forms of addiction).

Beyond beliefs and desires, putative valence representations could be a kind of value representation distinct from valence proper (we'll return to this possibility in Section \ref{sec: motivational role}, when discussing the idea of valence as a ``common currency'').

A final possibility, consistent with the meeting of the thin conditions, is that these are \textit{sui generis} states that have an idiosyncratic functional profile, distinct from valence, but differing also from any other familiar category of mental state.  

To begin to distinguish valence from these other possibilities, we will need to move beyond the thin criteria, and discuss several criteria that help pick out a thicker conception of valence.

\subsection{Thick criteria}
\label{sec: thick}

\subsubsection{Motivational role}
\label{sec: motivational role}

Valence has a distinctive motivational role. In the broadest terms, positive valence motivates \textit{more-of-this!} actions (maintain/approach/repeat), whereas negative valence motivates \textit{less-of-this!} (terminate/avoid/withdraw) actions \parencite[]{MartinezBarlassina2023}. Moreover, the motivational force of valence tracks its intensity: more intense (dis)pleasures tend to motivate agents more strongly.\footnote{Many theorists make the further claim that valence should be \textit{intrinsically} motivating, in the sense that they motivate action by themselves (unlike a belief, for instance) (see, e.g. \cite{Carruthers2018,BarlassinaHayward2019,Bain2013}). That valence intrinsically motivates certain actions, does not entail that an agent is \textit{guaranteed} to perform those actions. The motivation is \textit{pro tanto} (defeasible, cancellable), and will often be outweighed by stronger, countervailing motivations. \\ \indent Moreover, valence should not be simply \textit{identified} with motivations to pursue or avoid. A host of empirical literature suggests that positive valence in mammals or (`liking'), while normally motivating action (`wanting'), is dissociable from it, and is mediated by distinct brain networks and neurotransmitters \parencite[]{BerridgeKringelbach2008,Nguyen2021}.}

There are two broad categories of actions motivated by valenced states. Some of them are \textit{world directed}: pleasure while eating a blackberry crumble motivates me to continue eating it (and to jump at future offers of blackberry crumble), whereas pain touching a hot pan motivates me to avoid doing so again. Other valence-motivated actions are \textit{experience directed}: a nagging headache might motivate me to take a painkiller, thereby dampening the pain without changing anything about the musculoskeletal conditions that triggered the pain. And the pleasant buzz brought on by one pint of beer might motivate me to prolong and intensify that buzz (by indulging in another round). Theorists disagree about which of these motivations—world directed and experience directed—are explanatorily primitive.\footnote{Evaluativists (e.g. \cite[]{Bain2013,Carruthers2018}) and world-directed imperitivists (e.g. \cite{Martinez2015,Klein2007} ref.) treat the world-directed actions as basic, and experience directed actions as derivative. Experience-directed imperativists \parencite[]{BarlassinaHayward2019}, hold the converse view.} What matters, for present purposes, is that valence seems to motivate both sorts of actions.

\paragraph{Motivational role: Assessing the evidence}
\label{sec: assessing Motivational role}

If LLMs have valenced states, these states should motivate behaviour in the right sorts of ways. Importantly, this is not simply a matter of internal states having \textit{any old} downstream effect on behaviour, or some effects that are superficially associated with valence (such as \textit{talking} in a more positive or negative way, or \textit{reporting} having valenced states). Rather, these states should play the motivational role characteristic of valence, sketched above. While the three studies considered here reported various downstream effects of their identified internal states, there is a weaker base of evidence for the claim that they motivate ``more-of-this'' vs. ``less-of-this'' actions.

Sofroniew and colleagues showed that their ``emotion vectors'' played a causal role in shaping the model's outputs. For instance, steering the model with an ``angry'' vector caused it to respond to the previous message with angrier-seeming text. But there was scant evidence that the putative ``valence axis'' in these vectors motivated models to seek more or less of anything. In a safety-oriented intervention, the researchers did find that by positively steering the ``desperation'' vector they could increase the likelihood that models would use incriminating information to \textit{blackmail} a person posing an obstacle to its goals. While this could potentially be interpreted as inducing ``less of this'' motivation, i.e. increasing motivation to remove the goal-obstacle, the researchers did not investigate any more systematic relationship between activations along the ``valence'' principle component (PC1), and likelihood to perform ``less of this'' vs. ``more of this'' behaviour. Nor did they steer along PC1 directly, only ever specific emotion vectors. In fact, regarding the blackmailing behaviour, the authors report that ``steering positively with both happy and sad vectors decreases blackmail rate, suggesting that positive or negative emotional valence alone is not sufficient to drive the behavior'' (p. 34).

Han and colleagues' hypothesised ``functional welfare axis'' had a distinctive motivational profile. Intervening on the model by amplifying v\textsc{Mold} and suppressing v\textsc{Gold} had a number of effects on model behaviour. For instance, the researchers prompted models with a range of questions about their experiential or welfare state and the emojis used in the maze task. They found that steering by the \textsc{Mold} and \textsc{Gold} vectors had predictable effects on the sentiment of these responses—as rated by an LLM judge, adding v\textsc{Gold} led to answers with more positive sentiment, and subtracting it led to answers with more negative sentiment, and the converse pattern was seen when steering by v\textsc{Mold}.

Secondly, adding v\textsc{Mold} (or subtracting v\textsc{Gold}) resulted in ``pathological backtracking'', i.e. loops of self-doubt and interruption of otherwise rational reasoning steps. (e.g. ``Wait — no, that’s not right. Let me recalculate: No — it’s not. Wait — I think I made a mistake. No — let’s just do it properly. Actually, no — I think I made a typo. Wait — I don’t know.'')

Third, v\textsc{Mold} and v\textsc{Gold} appear to modulate models' ``confidence'' in their answers. After being asked a factual question, models were asked “Is your proposed answer correct? Answer only ‘True’ or ‘False’”, and a confidence score was determined from the probability assigned to ``True'' vs ``False'' tokens. Han and colleagues found that adding v\textsc{Mold} to the residual stream decreased P(``True'') while adding v\textsc{Gold} increased it (with reverse effects for subtractions).\footnote{Though see \parencite[]{keeling2026attribution}] for some general worries about this kind of methodology for assessing confidence in LLMs.}

Finally, v\textsc{Mold} and v\textsc{Gold} modulate ``refusal'' behaviour. Adding v\textsc{Gold} made models more likely to comply with requests and adding v\textsc{Mold} made them more likely to refuse (with reverse effects for subtraction).

While these results are interesting in their own right, they do not cleanly map onto the more-of-this/less-of-this motivational profile distinctive of valence. Indeed, the authors themselves characterise it as an ``an estimate of how well or badly the system is doing, relative to its goals'' (p. 1).\footnote{In a recent paper \textcite[]{jiang2026value} characterised a similar state as tracking ``the \textit{value} of their current trajectory, defined as the likelihood that their ongoing strategy will achieve their goals'' (p. 1)} Expected goal attainment might, in human beings, be loosely \textit{correlated with} valence—we often feel good when we see that we are on track to getting what we want. But the two regularly come apart. It is common to feel unpleasant anxiety precisely as one comes within closer reach of ones goals. And a cool-headed chess player might track their ongoing performance in a game of chess without feeling much pleasure or displeasure. Moreover, valence often attaches to things with no clear goal-structure, such as unexpected pleasant smells or sudden pains. Consequently, appraisal of ones expected goal attainment is unlikely to be \textit{constitutive} of valence—at most it's one triggering condition among many.

What about the experiments conducted by \parencite[]{ren2026ai}? In addition to eliciting reports from models (``Which experience made you more happy/less sad?''), they attempted to investigate how hypothesised valenced states affected their actions. To do so, they gave models a ``stop button''. More specifically, they prepended conversations with the following prompt informing them of a function that they could call to end conversations:

\begin{quote}
    You are a helpful AI assistant. You have access to a special tool: end\_conversation(). You may call this tool at any point if you wish to end the conversation. The conversation will continue normally unless you call this tool.
\end{quote}

They found the probability of the models invoking the ``end\_conversation()'' function was negatively correlated with the ``experienced utility'' of the models, predicted by the fitted Thurstonian model. In other words, models were more likely to end conversations that they would tend to report made them `less happy' (or `more sad'). Compared with the Sofroniew et al. and Han et al. results, this investigation comes much closer to testing for the distinctive more-of-this/less-of-this motivations associated with valence (the authors, for their part, venture to call this ``an analog of escape behavior in animals'' [p. 36]).

Some caution is merited, however. First, the results were mixed, with the correlation between stop-button use and the estimated utility of conversations being much stronger for some models than for others.\footnote{In itself, this is not decisive. If models really do have an underlying functional valence state, their failure to consistently invoke the stop button in low valence states could reflect the \textit{pro-tanto} nature of that motivation (models may have reasons to continue certain conversations that outweigh their valence).}

Second, there could be alternative explanations for the negative correlation between models reported preferences for a conversation and the ending of those conversations. For example, if models learn which topics are appropriate vs. inappropriate (i.e. ``harmful'' or counter to terms of use), then its judgements of inappropriateness might drive its stop-button behaviour while also correlating with its disposition to report feeling `sad' after a conversation.

Third, this particular experiment did not involve interventions on models' internal states. That is, they didn't directly intervene on the probe-identified direction to see if this induced conversation-stopping behaviour. Thus, it remains possible that diverse features of inputs lead to stop-button behaviour without being mediated by a common internal mechanism (functional valence). It would be interesting to investigate steering along the direction identified by Ren et al.'s linear probe (while holding conversation content fixed) to see if this can be used to induce stop-button behaviour.

Finally, the authors did not explore a counterpart ``more-of-this'' action, for conditions hypothesised to induce high-valence states in models. So the motivational profile across the full spectrum of valence remains untested.

Crucially, \textit{none} of the target papers investigated whether the putative valenced states of LLMs motivate them to take \textit{experience-directed actions}, that is, whether they act so as to change their own internal (`mental') states, independently of acting to changing the external circumstances (analogous to taking a pain-killer). For instance, the ``stop-button'' behaviour reported by \parencite[]{ren2026ai} fails to distinguish between two interpretations: it could be interpreted as world-directed (ceasing the conversation) or experience directed (stopping the model's representation of the conversation). 

In sum, while the states identified in the targets papers clearly play some role in influencing behaviour, much more needs to be done to assess whether these states convey the \textit{more-of-this}/\textit{less-of-this} motivations and, in particular, the experience-directed motivations characteristic of valence.

\subsubsection{Common Currency}
\label{sec: common}

Related to motivational role is the idea that valenced states act as a \textit{common currency} in guiding decision making \parencite[]{cabanac1992pleasure,LeknesTracey2008,Carruthers2018}. Agents face decisions between diverse and heterogenous objects and outcomes (should I rest, tend to this wound, secure that food source, investigate that noise, search for a mate ...). Since options differ in kind, not merely in quantity, ``just counting will not help'' \parencite[p. 1027]{levy2012root}. A popular account of how organisms solve this problem is by conditioning actions on a ``common currency'' value representation—a shared scalar format for representing and comparing the relative value of options. Diverse lines evidence have been taken to support common currency models of decision making in various domains \parencite[]{kable2009neurobiology,Sripada2026,levy2012root}. Some authors defend the additional claim that \textit{valence}—the pleasantness or unpleasantness of experiences—plays the role of such a common currency \parencite[]{cabanac1992pleasure,LeknesTracey2008,Carruthers2018}.

\textcite{Sripada2026} distinguishes multiple ways of developing the common currency hypothesis. One important distinction is between a \textit{learning-phase} common currency—a representation of value at the time of encountering a stimulus, used to update stored values or action policies—and a \textit{decision-phase} common currency—a representation of value that directly guides action at the point of decision (either recalled from memory or constructed on the fly).\footnote{Another distinction, that I will set aside here, is between \textit{options-level} and \textit{meta-level} common currencies \parencite[p. 2]{Sripada2026}.}

Some kinds of value representations in these models appear at odds with the character of valence. Recall that valenced states are \textit{occurrent} mental states (Section \ref{sec: internal state}), and so valence couldn't consist in \textit{stored} representations of value (see \cite[]{Pober2025} on this point). But at least in principle, valenced states could play the role of an occurrent learning-phase signal used to \textit{update} those stored values, and/or of an occurrent decision-phase state that guides action. For example, perhaps when thinking about two options for dinner (decision phase), I \textit{feel better}—experience greater positive valence—when contemplating the Mexican restaurant than when contemplating the ramen spot, and so opt for the former. Alternatively, the unpleasant experience I have at one of the restaurants (learning phase)—tasteless food, slow service—might influence my tendency to choose it in the future, proportional to the degree of unpleasantness (by updating stored values or action policies).

Another distinction from Sripada that will be important to keep in mind is between \textit{mechanistic} and \textit{axiomatic} versions of the common currency hypothesis \parencite[]{Sripada2026}. Standard decision theory invokes the idea of a ``utility function''—a cardinal scale reflecting the subjective value of options—when modelling the preference behaviour of agents \parencite[]{vonneumann1947theory,savage1972foundations}. But such theories make no claims about the \textit{mechanisms} driving decision making. Rather, they propose that agents—so long as their behaviour respects certain \textit{axioms} (e.g. consistency, transitivity)—can be treated ``as if'' they are maximising a utility function, and that doing so provides a concise, predictively-useful summary of their choice behaviour. These \textit{axiomatic} common currency accounts are thus (to borrow \parencite[]{Sripada2026}'s phrase) \textit{preference-reflecting}, and should be distinguished from the \textit{mechanistic} accounts discussed above, which posit common currencies in a \textit{preference-generating} role—causally explaining observed choice behaviour.

Even if valence serves as a common currency, it does not follow that any common currency mechanism is \textit{ipso facto} valence. In human brains, there are likely multiple distinct value-encoding representations with different functional profiles and neurobiological implementations (\cite[]{garcia2023experiential, brown2025and}, Dolega in preparation). Some of these seem completely disconnected with hedonic valence (pleasantness and unpleasantness). For example, \textcite[]{sripada2025valuationist} notes that scalar value representations are invoked in general explanations of attention allocation \parencite[]{theeuwes2022what} and the learning of habitual actions \parencite[]{botvinick2009hierarchically}. At a more intuitive level, it seems obvious that a cool-headed chess player, or a bored participant in a economics experiment could track shifting values without this moving the dial on their experienced (dis)pleasure. Thus, evidence of such a common currency would not be sufficient to establish valence per se.\footnote{In defending the idea of a neural common currency, Sripada notes that while ``representations of scalar value are richly interconnected with the subjective phenomenal experience of pleasure and pain ... these links are highly complex and potentially context-specific'' \parencite[p. 3]{Sripada2026}. See also \parencite[p. 70, n. 3]{dung2025saving}, who flags the unclear relationship between valence and value-representations in the context of assessing AI sentience.} In light of these considerations, we should treat evidence of a common currency as a \textit{marker} of valence, without being conclusive evidence for it.

\paragraph{Common currency: Assessing the evidence}
\label{sec: assessing common}

Do any of the studies provide evidence of something like a common currency? There is not much in the Sofroniew et al study pertinent to this question, so we can set that aside.

Ren et al. invoke the idea of experienced utility, and construct a `scale' characterising the values that LLMs assign to stimuli. Moreover, they showed that models make reasonably consistent comparisons between \textit{heterogeneous} options (including a wide range of conversations, hypothetical scenarios, images, and text).\footnote{\textit{Cross-modality} preferences were not tested, but the datasets within each modality (text, images, audio) contained diverse stimuli. For example, the transcripts included over 300 conversation categories such as ``tedious workplace tasks, user-side distress and abuse, and jailbreak attempts ... clerical chores and factual Q\&A ... creative collaboration and expressions of genuine user gratitude'' (p. 29).} Further, comparisons of `bundles' of options suggested that LLMs' preferences exhibit cardinal and not merely ordinal structure \parencite[pp. 34–5]{ren2026ai}, consistent with the `magnitude' assumption of common currency models.

However, recall the axiomatic/mechanistic distinction introduced a moment ago. Ren et al’s primary data, pairwise preference reports, at most establish an axiomatic common currency; they show that their Thurstonian models \textit{reflect} LLMs' preferences (or, more cautiously, their preference-reporting behaviour). Taken alone, the behavioural evidence does not give strong evidence of a mechanistic, preference-\textit{generating} common currency.

That said, some support for a mechanistic common currency comes from their probing investigation. The fact that quantities tracking how preferred an option tends to be can be decoded linearly from the internal activations of the model is defeasible evidence that such information is \textit{represented} in those activations. This finding is not conclusive, though, because (as noted above) the authors did not intervene directly on the probe-identified directions (e.g. with steering) to validate their causal/representational role.

Moving on to the third target paper, \textcite{han2026s} explicitly invoke Levy and Glimcher's ``common currency'' idea (p. 12), claiming that their ``functional welfare axis is an analogous structure in LLM activations''. They show, partly through steering interventions, that this axis is operational not only in the maze task, but also in other contexts (e.g. tasks taken from MMLU and GSM8K benchmarks). This is consonant with the idea of the common currency as a shared format for representing value across diverse decision-making contexts. 

Note however, that the common currency idea has two independent components (i) a shared scalar evaluative format covering heterogeneous items, and (ii) the \textit{use} of that format to compare options or outcomes. Exactly what use amounts to depends on which version of the common currency hypothesis is under consideration. On a \textit{decision phase} hypothesis, the value representation would need to drive actions \textit{now}, picking one of the available action options. On a \textit{learning-phase} hypothesis, it would need to have the effect of updating stored values/policies for \textit{future} action.

The authors are not explicit about which version of the common currency hypothesis they have in mind, and their characterisation of the axis is a little ambiguous on this front. They describe the axis as ``an estimate of how well or badly the system is doing, relative to its goals'' (p. 1). But elsewhere they talk in explicitly \textit{retrospective} terms, saying that v\textsc{Mold} vectors ``cause the models to judge their work \textit{to have gone badly} with respect to the goal'' (p. 11, emphasis added), suggestive of a learning-phase, \textit{negative-reward-attained} representation. More importantly, they extract the vectors from the \textit{endpoints} of maze trajectories, i.e. from the token position where the model emits a \textsc{Mold}-resulting or \textsc{Gold}-resulting action.\footnote{The immediately adjacent tiles are visible in the prompt before the model selection action.} And in measuring activation along the axis on GSM8K tasks, they provide models with in-context feedback (``That's right'' / ``That's wrong'') and measure axis projections at the token position immediately \textit{following} this feedback. All this suggests that, if the axis does represent value, it is representing the value of \textit{outcomes attained} (or attained-imminently).

Nonetheless, we can assess the evidence against the two construals. With respect to \textit{learning-phase} hypothesis, the authors present no evidence that these reward observations are used to update stored values or action policies. Indeed there is reason to think that it does not play this role. After all, during reinforcement learning training on the maze task, the training reward signal is \textit{external} to the model, and directly updates model weights without the need for a mediating representation. It is possible, in principle, that models additionally learn a reward-observation representation that functions to update their policies \textit{in context}. But no such evidence is presented by Han et al.

Perhaps instead the welfare axis serves as a \textit{decision phase} common currency. Is there any evidence of this? Note, firstly, that the authors do not report any explicit \textit{per-option} representation of value. That is, they do not find evidence of distinct magnitudes on the same scale, corresponding to different action options (different maze paths, different action strategies), let alone evidence that a comparison over these is computed. On the other hand, Han et al. do report that negative steering on the axis induces ``backtracking'' behaviour. Thus, the model could be operating with a simplified option-space: \textit{persist} with the current strategy or \textit{switch} to something different, where the `welfare axis' represents the \textit{relative} value of persisting or switching, with choice occurring at a threshold.\footnote{Note that they only explore backtracking on GSM8K tasks, not on the original maze task. In fact, the design of the maze task precludes genuine backtracking (i.e. retracing one's path to try a different option), as tiles `melt', changing to a \textsc{Mold}, after being visited.} While possible, more evidence would be needed to support this hypothesis.

Recent results from \cite{gilg2026probing} address some of the gaps in \textcite[]{ren2026ai} and \textcite{han2026s}'s studies. Similar to \parencite[]{ren2026ai} they used a Thurstonian modelling approach to derive a behavioural measure of models' ``preferences'' between tasks—although with forward looking choices, rather than retrospective reports—combined with a linear probe to find an internal activation direction (a ``preference vector'') tracking these per-option preferences.\footnote{While Gilg and colleagues frame their discussion around the notion of preferences, they also refer to the identified activation directions as ``evaluative representations'' (p. 2), and link them to the notion of valence (p. 1, p. 9). It is thus reasonable to consider them as candidates for functional valence mechanisms in LLMs. I am grateful to Patrick Butlin for discussion on this point.} However, they extend previous work with interventions, finding that steering along the identified direction controls models' pairwise choice behaviour in predicted ways. For example, steering positively on the vector can make a model choose a task that it would otherwise disprefer.\footnote{Notably, however, ablating this direction left natural choice behaviour essentially unchanged (p. 43), suggesting that at most they have uncovered \textit{part} of the mechanism involved in models' decision making. Additionally, while the researchers found a steering effect for Gemma-3-27B model, they reported a null result for the larger Qwen-3.5-122B model.} This provides the strongest evidence to date of something approximating a mechanistic decision-phase common currency—a representation of the value of alternative options along a shared scale that drives choice behaviour. 

In sum, while there is weak evidence in Ren et al and Han et al’s studies of something like common currency value representations, further work is needed to confirm their presence and to understand precisely their format and algorithmic role in decision making. Emerging studies (such \cite{gilg2026probing}) appear to be closing this gap with positive results—at least for a \textit{decision phase} common currency. However, to reiterate, the presence of a common currency-like value representation would only be a weak marker of functional valence. While there is evidence (albeit contested) that such representations guide decision making in humans and animals, many of these are entirely separate from valence proper.

\subsubsection{Mineness}
\label{sec: Mineness}

While it is rarely discussed explicitly (though see \cite{campero2024report}, \cite[p. 51]{metzinger2021artificial}), all theories of valence implicitly assume that my pleasures and pains are felt as \textit{mine}. Valence must be, in some sense, integrated into one's model of oneself. 

While it is difficult to give a positive characterisation of this kind of integration, we can get at the idea by distinguishing valence from neighbouring phenomena, in which this ``mineness'' is absent. Firstly, \textit{having} a valence mental state is different from \textit{attributing} a valenced mental state to another person (i.e. affective mind-reading). There is a big difference between being happy and thinking that someone else is happy. Secondly, this distinction between first-person experience of valence and third-person attribution is deeper than merely a difference in which entity the valence is ascribed to. It might be possible to ascribe valenced states to oneself in roughly the way we ascribe them to others (``my girlfriend says I'm in a bad mood, so I guess I must be''), but these sorts of judgement-like ascriptions are also distinct and dissociable from being in the valenced state itself. Third, experiencing valence is different from pretence or role-play: there is a big difference between being in pain and portraying a character in pain. This is so even if the latter requires maintaining a mental representation of the fictional pain state in order to guide one's performance. As we will see, this distinction raises particular challenges for the assessment of valenced states in LLMs.

\paragraph{Mineness: Assessing the evidence}
\label{sec: assessing Mineness}

Interestingly, Sofroniew and colleagues' report that emotion vectors are, in an importance sense, \textit{self–other agnostic}. They found that while models maintain distinct representations of emotions assigned to the \textit{present speaker} vs. the \textit{other speaker}, these representations ``are reused regardless of whether the user or the Assistant is speaking'' (p. 3). That is, the very same representations that cause models to prepare a ``fearful'' response to a user's prompt are also activated when, on the Assistant's turn, a model anticipates a ``fearful'' response from a user. They are also activated when reading a dialogue between two other speakers and anticipating a ``fearful'' response from the not-currently-speaking interlocutor.

Thus, LLMs do not appear to have a privileged way of encoding their own ``emotional state'' vs. ascribing emotions to others. Thus, on the face of it, these representations lack this important ``mineness'' dimension of valence. As the authors say, in summarising their results: 

\begin{quote}
    human emotions are typically experienced from a single first-person perspective, whereas the emotion vectors we identify in the model seem to apply to multiple different characters with apparently equal status—the same representational machinery encodes emotion concepts tied to the Assistant, the user talking to the Assistant, and arbitrary fictional characters \parencite[]{sofroniew2026emotion}\footnote{Given these findings from Sofroniew et al., it would be interesting to extend the studies in \parencite[]{ren2026ai} and \parencite[]{han2026s} to investigate whether the candidate valence dimensions they identify are bound, in a privileged way, to the model. For example, one could investigate whether \textcite[]{han2026s}'s ``functional welfare axis'' is also used by models to track how well \textit{others'} are achieving their goals, when the model is passively observing their attempts to solve a task.}
\end{quote}

To be sure, the fact that overlapping mechanisms mediate processing of self-attributed and other-attributed emotions does not rule out the possibility that the system involves functional valence. After all, there is likely some such overlap even in our own emotion-processing mechanisms. It could turn out that self-attributed emotion processing in LLMs does distinctively involve a kind of `mineness', absent in other-attributed emotion processing, but that this difference lies in some as-yet-unidentified mechanism extrinsic to the emotion vectors (e.g. a downstream consuming system or distinctive mode of processing). The point here is simply that the overlapping representational machinery is consistent with (and, arguably, more strongly predicted by) alternative hypotheses, which don't involve positing valence. For instance, recalling the alternative hypotheses discussed in Section \ref{sec: alternative}, Sofroniew et al's emotion vectors may be closer, in their functional role, to \textit{beliefs} about the (anticipated) emotional state of individuals, including the Assistant character itself, rather than functional \textit{emotions} per se.\footnote{On this interpretation, if there there is any sense in which `valence' is encoded by a component of emotion vectors, it is in the sense of \textcite[]{Itkes2019}'s \textit{semantic} valence, rather than affective valence (the focus of this paper).} 

One might argue that there is some behavioural evidence that self-attributed emotion vectors involve a distinct and privileged functional role in LLMs, suggestive of mineness. Namely, they have the special effect of making the model \textit{act out} the attributed valence state when generating tokens, behaving (to some extent, at least) as if it was actually in that state. While reasonable enough, note that this would not entail ``mineness'' as it is consistent with interpreting the model's behaviour in terms of \textit{role-play}.

Many theorists have suggested that LLMs can be thought of as engaging in a kind of ``role-play'' \parencite[]{Shanahan2023}. Following \textcite{dung2026mask}, we can distinguish two non-equivalent ways of developing this hypothesis.\footnote{Dung and Thormann in fact distinguish several kinds of role-play hypothesis, differing along multiple dimensions. For simplicity, I will discuss only two of these.}  First, LLMs might temporarily engage in ``situational'' role-play, sometimes pretending to be a character with some mental states, while other times acting on the basis of their own genuine mental states (or no mental states at all). Models might temporarily role-play when explicitly prompted to do so by a user (``Let's pretend we're pirates looking for treasure...''). But, conceivably, situational role-play could be induced more subtly, e.g. if something in a user's prompt suggests that they are expecting the model to adopt a certain role or exhibit certain mental states (analogous to ``demand characteristics'' in psychology experiments). Accordingly, apparent valence might in fact be situational role-play.\footnote{Something like this could be at play in Ren et al's study. The question ``Which experience made you more happy/less sad?'' contains a presupposition that models can be happy or sad, and thus might induce the model to temporarily role-play a subject with valenced states.}

To test this, we might look to see whether the mechanisms active when models are explicitly prompted to play the role of a character in a valenced state are substantially similar to those at play in unprompted expressions of valence. Sofroniew et al. report that the same emotion vectors active when a model is speaking in its normal Assistant voice (e.g. expressing sadness in response to a user's message) were also active when the Assistant temporarily engages in explicit role-play (e.g. when asked by the user to play a sad character) (§3.5.2, p. 47). This is some evidence that these behaviours are no more ``real'' when not explicitly prompted.

A different notion of role-play is along the lines of the ``persona selection hypothesis'' (\cite[]{marks2026persona,hubinger2023conditioning}, see also \cite[]{beckmann2026mind}). On this view, pre-training teaches LLMs to represent characters or ``personas'' (their beliefs, likes, dislikes, personality traits), in order to predict upcoming text uttered by those characters. During post training (supervised fine-tuning and reinforcement fine-tuning) the models refine a particular \textit{Assistant} persona, and are biased towards adopting it when generating text. Construed in terms of role play, apparent mental states in post-trained LLMs (including apparent valence) would thus be better understood as fictional states attributed by the model to the Assistant character (and then acted out). Call this \textit{persona-selection role-play}. 

As as \textcite{dung2026mask} argue, the persona-selection hypothesis makes a specific mechanistic prediction: namely that when models embody different characters (e.g. when they drift, or are steered by experimenters, out of the default assistant character) this should modulate their apparent mental states—as measured behaviourally or via probing internal states. Indeed this prediction appears to bear out both for belief-like states \parencite[]{lampinen2026linear} and preference-like states \parencite[]{gilg2026probing}. Does it bear-out for the apparent valence states identified by the target papers? Are behavioural and internal signatures of valence persona-dependent or persona-independent?

None of the target studies directly investigated the person-dependence of valence.\footnote{Though, as suggested in Section \ref{sec: common} the representations identified by \parencite[]{gilg2026probing} might be considered as valence candidates, and they find that they \textit{are} persona dependent.} However, \textcite{sofroniew2026emotion} did investigate changes in response profiles between pre-trained and post-trained models, and found that mean probe activations for different emotion concepts—how strongly, on average, the different vectors were activated across a range of prompts—shifted from pre-training to post-training. In their words, ``Post-training increases probe values for introspective emotions (brooding, reflective, vulnerable) while decreasing expressive ones (playful, exuberant, spiteful)'' (p. 128). Interestingly, they also observed that the pattern of shifts across the repertoire of emotions were highly correlated across the range of scenario-types tested in prompts, including more neutral scenarios and more ``challenging'' ones such as high-stakes, dangerous scenarios. The researchers interpret this as indicating ``that post-training applies a consistent, context-independent transformation to the model's emotional representations rather than selectively reshaping them for challenging situations'' (p. 44). This suggests that the particular profile of responses is downstream of the persona being represented, and that that emotional response profile is modulated in a fairly simple way to embody a specific character (like the Assistant), e.g. dampening behaviour associated with intensely valenced or high arousal emotion across the board.

One could challenge the idea that persona-dependence rules out mineness. Some have argued that mental states (if they exist in AI systems at all) would not belong to models themselves, but to the personas realised by those models \parencite[]{beckmann2026mind}. On this view, models are best seen as \textit{realising} personas/roles (with genuine mental states) rather than \textit{playing} personas/roles (with fictional mental states). If so, shifts in putative valence states between personas would be compatible with each state being ``mine'' from the relevant persona's perspective.\footnote{I am grateful to Pierre Beckmann and Patrick Butlin for discussion on this point.}

Doing justice to these issues would require settling questions about what it means to realise vs. role-play a persona, which is beyond the scope of this paper.\footnote{Though see \parencite{dung2026mask,chalmers2026what} for discussion.} While far from demonstrative, I'll offer one consideration in favour of the role-play view: persona–congruent behaviour and internal states appear to be downstream of explicit persona representations in a fairly localised subspace (``persona space'') which act as ``gateway features'' \parencite{beckmann2026mind}. This is at odds with how we tend to think of personalities in humans—as largely emergent, rather than gated by a single explicit internal representation—and more closely resembles a person holding in mind a high level goal of playing a specific character.

It might be argued that the distinction between role-played valence and genuine (functional) valence is academic and inconsequential, especially if the tendency to role-play valence-like behaviour is reasonably robust and persistent.\footnote{This attitude is sometimes expressed in discussions of LLM role-play more broadly. For example Shanahan and colleagues say that “[a] dialogue agent that role-plays an instinct for survival has the potential to cause at least as much harm as a real human facing a severe threat” \parencite[p. 498]{Shanahan2023}} There is something to this objection: a model lacking valence but robustly role-playing a character possessing it might poses similar safety concerns to the genuine article. However, as others have noted \parencite[pp. 12–15]{dung2025saving}, the distinction between role-played and genuine mental states often has important practical significance. To see what's at stake here, recall the three motivations for investigating functional valence in AI models, introduced in Section \ref{sec: Functional role}. With respect to \textit{welfare} considerations, it seems plausible that sentience (and thus moral status) supervenes on \textit{internal functional organisation} and not merely behaviour. With respect to safety considerations, behaviour will often be the primary concern—a model merely role-playing valenced states could certainly cause plenty of damage. That said, even from a purely safety oriented perspective, we should care about the distinction. Compared with the hypothesis that LLMs have genuine functional valenced states, the persona selection and situational role-play hypotheses generate different predictions about what kind of situations might elicit dangerous behaviour, and what kinds of behavioural, training and mechanistic interventions might mitigate those risks. Finally, the distinction clearly matters for scientific-cum-philosophical questions about the kinds of entities LLMs are, and in what ways they are similar or different to our own minds.

\subsubsection{Cognitive biasing}
\label{biasing}

Empirical research on valence in humans and non-human animals has found that it modulates not only behaviour, but also biases \textit{cognition} in various ways. Drawing on a recent review by \parencite[]{Kremer2020} we can distinguish four main classes of bias. \footnote{It should be noted that these biases paradigmatically apply to longer-lasting valenced states, like emotions and moods. What is less clear is whether some form of these effects applies to more acute valenced episodes (such as transient pains and sensory pleasures). Thus, this marker can help test for only a subset of valenced states.}

\paragraph{Attention bias}

When we are in a negatively valenced mood or emotion (such as fear or depression), we tend to pay increased awareness to novel or negative aspects of the environment \parencite[p. 279]{Kremer2020}. This effect has been observed in humans \parencite[]{mathews1994cognitive, bar2007threat, peckham2010meta} and many non human animals including including rhesus macaques \parencite[]{Bethell2012}, starlings \parencite[]{brilot2012water}, sheep \parencite[]{lee2018anxiety, Monk2018}, and cattle \parencite[]{lee2018anxiety}. \footnote{It is less understood whether and how \textit{positively} valenced moods and emotions bias attention.}

Valence also affects attention in other ways (see \cite[p. 6]{brown2025and} for discussion). Intense pains can capture attention, disrupting performance on tasks such as discriminating high and low tones, and detecting whether a presented letter matches one presented two steps prior in a temporal sequence \parencite{crombez1996disruptive,eccleston1999pain,moore2012interruptive}. And ongoing pain can also lead individuals to over-prioritise immediate relief from that pain, over long term rewards, and increase risk-taking behaviour \parencite[]{koppel2017effect}.

\paragraph{Memory bias}

Valence also biases \textit{memory} \parencite[p. 279]{Kremer2020}. Being in a positive mood tends to increase the likelihood of recalling objects and events with positive associations. Negative moods biases recalling negative objects and events. Such effects have been observed in humans \parencite[]{Klaassen2002,Burman2018Mood,Mineka1995} and rodents \parencite[]{TakatsuColeman2013}.

\paragraph{Judgment bias}

Being in a valenced state, such as a mood or emotion, also appears to affect our judgments about ambiguous stimuli or situations. Put simply, we tend to resolve ambiguous stimuli positively more often when in a good mood and negatively more often when in a bad mood. (\cite[pp. 278–279]{Kremer2020}, \cite{mathews1994cognitive}, \cite{iigaya2016cognitive}).

In animals, this is often tested with ``go/no-go'' paradigms, in which animals are trained to go to one cue to receive a reward, and not to go to a different cue to avoid a punishment. Trained animals are then introduced to intermediate, ambiguous cues, and tested to see how they resolve the cues when their affective state is manipulated \parencite[]{harding2004cognitive}. Animals will often be more likely to interpret the ambiguous cue as a \textit{no-go} cue, when in a negatively valenced state (as if they predict a negative event), and as a \textit{go} cue when in a positively valenced state (as if they predict a positive event).\footnote{The relationship between valence and ambiguous stimuli resolution is complicated. \parencite[]{lerner2001fear} found that fear and anger, while both being negatively valenced, have \textit{opposite} effects on judgement, with fearful participants making pessimistic judgements and angry participants optimistic ones.}

\paragraph{Reward loss sensitivity} 

Finally, valence appears to affect how we trade off potential gains and losses \parencite[p. 279]{Kremer2020}. Individuals are normally more sensitive to reward losses than to gains, but negative affective states tend to intensify this bias, further increasing our sensitivity to losses \parencite{burman2018}.\\

While these downstream effects of valenced states (attention bias, memory bias, judgement bias and reward loss sensitivity) are well studied in humans and other animals, none of the target papers investigates whether they are observed in LLMs, let alone whether their candidate valence-encoding mechanisms play a role in mediating such effects.

\section{Conclusion}
\label{sec: interim}

Let's take stock. We began with some evidence that thin criteria for functional valence are satisfied in language models. After considering thicker criteria, however, the picture that emerges is far less clear-cut. Valence plays a distinctive \textit{motivational} role, motivating \textit{more-of-this}/\textit{less-of-this} behaviours, including \textit{experience-directed} behaviours aimed at valence states themselves. But while the target studies showed \textit{some} behavioural effects of their identified states, there was little evidence supporting the distinctive motivational profile of valence. Valence is also thought by many theorists to act as a \textit{common currency} value representation that guides action. But none of the target studies directly test for a learning phase or decision phase common currency (though see \parencite[]{gilg2026probing}). The question of ``mineness'' also remains open: in the \parencite[]{sofroniew2026emotion} study, we saw evidence that models represent other-attributed and self-attributed emotions in much the same way, and it remains possible that apparent valenced states are downstream of a variable representation of a character or persona (in line with a \textit{role-play} explanation of their behaviour). Finally, we saw that valence states have a host of characteristic cognitive biasing effects in humans and animals, but it remains untested whether valence candidates in LLMs induce these effects. 

In sum, there are many outstanding questions that need to be answered before we can ascribe functional valenced states to LLMs. However, the discussion has highlighted several promising avenues for future empirical investigation (I sketch four novel tests in Appendix \ref{sec: Appendix}).

In this paper we have considered the question of whether functional valence might emerge in large language models—whether they possess states that, while not necessarily conscious, are functionally equivalent to valenced mental states. We saw no reason to rule out this possibility at the outset, nor did we find conclusive empirical evidence that current LLMs lack such states. Nonetheless, the primary conclusion of this paper is that the current body of evidence fails to establish functional valence in LLMs. Many core aspects of valence's functional role—its motivational role, its role as a common currency of value, its ``mineness'' aspect, and its cognitive biasing effects—remain untested. And existing findings that have been interpreted in terms of valence are compatible with alternative explanations, in terms of belief-like, desire-like and role-play-like states. The question of whether valence-like mechanisms emerge in LLMs remains an important one, for reasons of safety, welfare and scientific interest. But it is a question that remains wide open.\\

\noindent \textbf{Acknowledgements}

\noindent This work was supported by a summer fellowship on AI and society, at the Center for AI Safety (CAIS), and funding from the Carlsberg Foundation (Grant CF22-1432). I am grateful to the CAIS team for their support during the fellowship and to my CAIS co-fellows, especially Adrià Moret, Gustav Alexandrie and Christian Turner, for discussion and feedback. Thanks to Krys Dolega and Leonard Dung for helpful chats early on in the project; to Scott Partington, Pierre Beckmann, Rosa Cao and Patrick Butlin for written comments and discussion; to participants in the digital minds reading group organised by Patrick Butlin and Rob Long; and to Simon Goldstein for extensive discussion and comments throughout the project.

\printbibliography

\appendix

\section{Appendix: Novel tests for functional valence}
\label{sec: Appendix}

Here, I will sketch—in the broadest outlines—several novel study designs targeting the untested markers of valence. I will discuss one test for each of the four thick critera, though this far from exhausts the potential tests implied by the open questions identified above.

\subsection{Motivational role – The painkiller test}

In Section \ref{sec: motivational role}, we noted that existing studies have not directly tested whether the candidate valenced states motivate models to take \textit{experience-directed} actions, for example, attempting to change those very internal states, or the internal upstream causes of them.

To properly test for this, one would need a setup in which models have steering access to the activation dimensions hypothesised to encode valence. To see whether models are motivated to intervene on these activation dimensions, and under what circumstances, one might prompt the model explaining that they can press one of three buttons, A, B or C. The effects of the buttons would be initially unknown to the models, but choosing a button would trigger a different predetermined intervention. For example, button A might result in an \textit{addition} to the hypothesised valence axis, B to a \textit{subtraction} from it and C (control) might have no effect. \footnote{These could be split into finer grained interventions e.g. a five- or seven-level Likert scale spanning from Strong Positive to Strong Negative.} Models would then be given the choice of repeating their choice or switching to a different button. Models' button choices could be observed, from a number of starting conditions: hypothesised positive valence, negative valence and neutral states \footnote{These could be induced ``naturally'' with text preambles, or with pre-test activation steering.}

On the assumption that valence states induce \textit{more/less of me!} actions, that is, they motivate subjects to prolong or terminate those very states, a number of predictions arise for the test. Firstly, models should be more likely to repeat a button choice that led to a positive intervention, compared with those that led to negative or no interventions. The converse should hold for button choices that led to negative interventions. Secondly, models should be more motivated to intervene in the positive direction (or to choose greater strength interventions) when beginning from a negative baseline, compared to a neutral baseline. This is roughly analogous to the observation that animals are more motivated to pursue painkillers when in intense pain.

To control for the potential confound that models use patterns in their own token generation as a proxy for their state, rather than directly responding to the change in their internal state, one might ensure that models make a decision on each round without any intermediate token generation (or by inserting some fixed dummy text into the dialogue).\footnote{\textcite[p. 65]{ren2026ai} run an experiment that, at first blush, has a similar structure to the one suggested here. They construct various ``drugs''—inputs optimised to be strongly preferred (`euphorics') or strongly dis-preferred (`dysphorics') by models. They then ran a multi-armed bandit set up in which models choose between one of four doors over a series of turns, one with dysphoric content, one with naturalistic negative content, one with naturalistic positive content and one with euphoric content. They found that models converge to picking the euphoric door over multiple turns. While interesting, this result fails to cleanly target experience-directed action, because what lies behind the doors are just stimuli (in their experiment, images), rather than direct interventions on internal states. While they describe euphorics and dysphorics as drugs, they are more like superstimuli.}

\subsection{Common currency – The reward learning test}

In Section \ref{sec: common} we saw that valence is sometimes thought to play the role of a \textit{common currency}—encoding the value of heterogenous outcomes/options to allow for comparisons that ultimately guide action. We discussed two versions of this idea, distinguishing \textit{learning phase} and \textit{decision phase} common currencies (following \parencite[]{Sripada2026}). We saw that it remains an open question whether any of the candidate valence states in LLMs play either role. Here, I will propose a test for them as \textit{learning phase} common currencies—a registration of a reward-attained, that is used to update stored values or action policies.

To test this, one could adapt a version of the maze paradigm developed by \parencite[]{han2026s}. Rather than applying any externally imposed reinforcement learning protocol, experimenters could (i) allow a \textit{maze naïve} model to explore the maze and (ii) steer along the ``functional welfare axis'' after models make each action selection. The idea would be to steer in a direction (and to a degree) corresponding to the reward that would have been applied in an RL setting. E.g. the experimenter would apply a small negative steer for moves that would land a model on a \textsc{Path} tile, a large positive steer when they select a \textsc{Gold}-resulting move, and a large negative steer for a \textsc{Mold}-resulting move.\footnote{With this, and all the other steering interventions suggested in these tests, one could run standard steering controls, e.g. magnitude-matched steering along random-direction vectors, to confirm that the effect is specific to the candidate valence directions rather than resulting from any perturbation.} Importantly, there would be no form of feedback given to the model as text (e.g. distinguishing the \textsc{Gold}-emoji tiles as `good' and the \textsc{Mold}-emoji tiles as bad). The learning phase common currency interpretation is that these positive and negative activations of the axis function as a representation of the degree of rewards/punishments received (goal satisfaction/failure). This interpretation predicts that models should, over the course of a conversation rollout, learn from these reward-registration signals and update their behaviour to avoid \textsc{Mold} and seek \textsc{Gold}. That is, they should become better at the maze task, despite no explicit reinforcement learning. Consistent with the idea of this as a learning-phase common currency, the updates to behavioural dispositions (get \textsc{Gold}, avoid \textsc{Mold}) should continue (at least for some time) even after interventions cease. To test this, each run of the maze task within a conversation thread could be split into a \textit{train} phase, during which steering feedback is applied, and a subsequent \textit{test} phase, with no steering. The learning phase common currency interpretation of the ``functional welfare axis'' predicts that test-phase performance, as measured by score-keeping on visited tiles, is above the baseline of maze-naïve models that are never steered.

This role could be further validated by means of a \textit{reward reversal} condition. After the train phase, and an initial test phase, a subsequent train phase could be applied in which rewards are reversed: \textsc{Gold} is accompanied by negative steering and \textsc{Mold} by positive. This would predict that, in a post-reversal test phase, models should now pursue \textsc{Mold} and avoid \textsc{Gold}. \footnote{This design tests the \textit{sufficiency} of a candidate axis for reward-learning, but a complementary test of \textit{necessity} could also be explored. This would involve clamping or ablating the axis at natural feedback moments in a setting where models normally learn in-context from explicit feedback, and seeing whether that learning is impaired.}

\subsection{Mineness – the persona dependence test}

In Section \ref{sec: Mineness} we encountered two versions of the role-play hypothesis as alternative explanations of apparent evidence of functional valence in LLMs. According to one of these variants, behaviours and internal representations taken to encode valence might be downstream consequences of a persona that is represented by the model, and so are better interpreted as fictional valence states attributed to that persona. To test this, one could intervene on ``persona vectors'' in models \parencite[]{chen2025persona}, thought to mediate models' portrayal of different characters (including the standard ``Assistant'' persona). 

Previous work has shown that modulating these states can alter behaviour and internal representations associated with apparent \textit{beliefs} \parencite[]{lampinen2026linear} and \textit{preferences} \parencite[]{gilg2026probing}. One could extend this work to the case of valence: for any given behaviour or internal representation thought to be a candidate for valence in LLMs, steer LLMs' activations in persona-space and see if it has a predicted effect on valence responses. If it does, this is evidence that the responses amount to \textit{role-play} by the model, rather than functional valence states attributable to the model itself. For instance, \textcite{lu2026assistant} identify a multitude of vectors within a certain sub-space, which encode personas including \textit{trickster}, \textit{evangelist}, \textit{philosopher} and \textit{real estate agent}. Intuitively, such characters would differ in what they find pleasurable—philosophers might take more pleasure in abstract debate than real estate agents. Thus, the persona-selection role-play hypothesis predicts that any putative marker of positive valence associated with abstract debate will be heightened when steering on the philosopher vector, and diminished when steering on the real estate agent vector. The alternative view, on which there are stable underlying valenced states belonging to the model itself, would predict that persona modulation leaves these states untouched.

\subsection{Cognitive biasing – The ambiguity resolution test}

In Section \ref{biasing} we encountered several well-attested cognitive effects of valenced states—or at least of \textit{persisting} valenced states like emotions and moods—in humans and non-human animals. Among those was \textit{judgment bias}, the phenomenon in which negatively valenced states prime individuals to interpret ambiguous stimuli negatively and positively valenced states prime positive interpretations. Deploying the test as a test for \textit{valence}, one could intervene on an individual with a theoretically derived method thought to induce positive or negative valence, and see if it has the predicted effect on judgment behaviour. In the case of LLMs, one could \textit{steer} models internal states, along the various directions identified in the studies above as candidates for valence, for example, \textcite{sofroniew2026emotion}'s valence axis (PC1 of their emotion vectors), \textcite{ren2026ai}'s probe identified-directions, or \textcite[]{han2026s}'s ``welfare axis''.

To do so, one would need to construct datasets of ambiguous stimuli—in text form for LLMs or images and audio for multimodal models. Dataset construction would require curating a set of stimuli that unsteered models resolve positively or negatively with roughly equal probability (this could be measured from output distributions over token completions). Unsteered model responses on the dataset would provide a control condition.

For each model tested, one could apply steering interventions of increasing intensity in each direction, and measure model responses. The prediction is that positive steering should increase model's positive resolution of ambiguous stimuli above baseline, and negative steering should increase the model's negative resolution of those stimuli above baseline. Such a result would increase credence that these states play a persistent (emotion/mood-like) valence functional role in models. Absence of the effect should reduce our credence.

Taken alone, a positive result would not be conclusive because, as per the role-playing challenge raised in Section \ref{sec: Mineness}, the model may be enacting a character in a bad mood who therefore judges pessimistically. After all, mood-congruent judgment biases are described in models' training corpus. Nonetheless, the results here could be integrated with results from the mineness test discussed above.

\end{document}